\documentclass[runningheads]{llncs}
\usepackage{graphicx}
\usepackage{comment}
\usepackage{amsmath,amssymb} 
\usepackage{color}

\usepackage[width=122mm,left=12mm,paperwidth=146mm,height=193mm,top=12mm,paperheight=217mm]{geometry}

\usepackage[pagebackref=true,breaklinks=true,letterpaper=true,colorlinks,bookmarks=false,citecolor=blue,linkcolor=blue]{hyperref}
\usepackage{url}
\usepackage{paralist}
\usepackage{subfig}
\usepackage{booktabs}
\usepackage{makecell}
\usepackage{paralist}
\usepackage[ruled,lined]{algorithm2e}

\def\eg{e.g.,~}               
\def\ie{i.e.,~}               
\def\vs{vs.}                 

\newlength\paramargin
\newlength\figmargin
\newlength\subfigmargin
\newlength\secmargin
\newlength\subsecmargin
\newlength\tabmargin
\newlength\eqmargin

\newcommand{\Paragraph}[1]
{\vspace{2mm} \noindent \textbf{#1}}

\newcommand{\figref}[1]{Figure~\ref{fig:#1}} 
\newcommand{\tabref}[1]{Table~\ref{tab:#1}}
\newcommand{\eqnref}[1]{\eqref{eq:#1}}

\long\def\ignorethis#1{}

\newcommand{\tb}[1]{\textbf{#1}}

\begin{document}

\newcommand{\lu}[1]{{\color{red}{(Lu\@: #1)}}}
\newcommand{\hyt}[1]{{\color{blue}{(HungYu\@: #1)}}}
\makeatletter
\newcommand{\printfnsymbol}[1]{%
  \textsuperscript{\@fnsymbol{#1}}%
}
\makeatother

\pagestyle{headings}
\mainmatter
\def\ECCVSubNumber{524}  

\title{RetrieveGAN: Image Synthesis via Differentiable Patch Retrieval}

\titlerunning{RetrieveGAN: Image Synthesis via Differentiable Patch Retrieval}
%
\author{Hung-Yu Tseng\thanks{Equal contribution. Work done during their internships at Google Research.}$^2$,
Hsin-Ying Lee\printfnsymbol{1}$^2$,
Lu Jiang$^1$,
Ming-Hsuan Yang$^{1,2,3}$,\\
Weilong Yang$^1$
}
\authorrunning{H.-Y. Tseng et al.}
%
\institute{$^1$Google Research\hspace{14pt}$^2$University of California, Merced\hspace{14pt}$^3$Yonsei University}

\maketitle

\begin{figure}[th]
    \centering
    \includegraphics[width=0.95\linewidth]{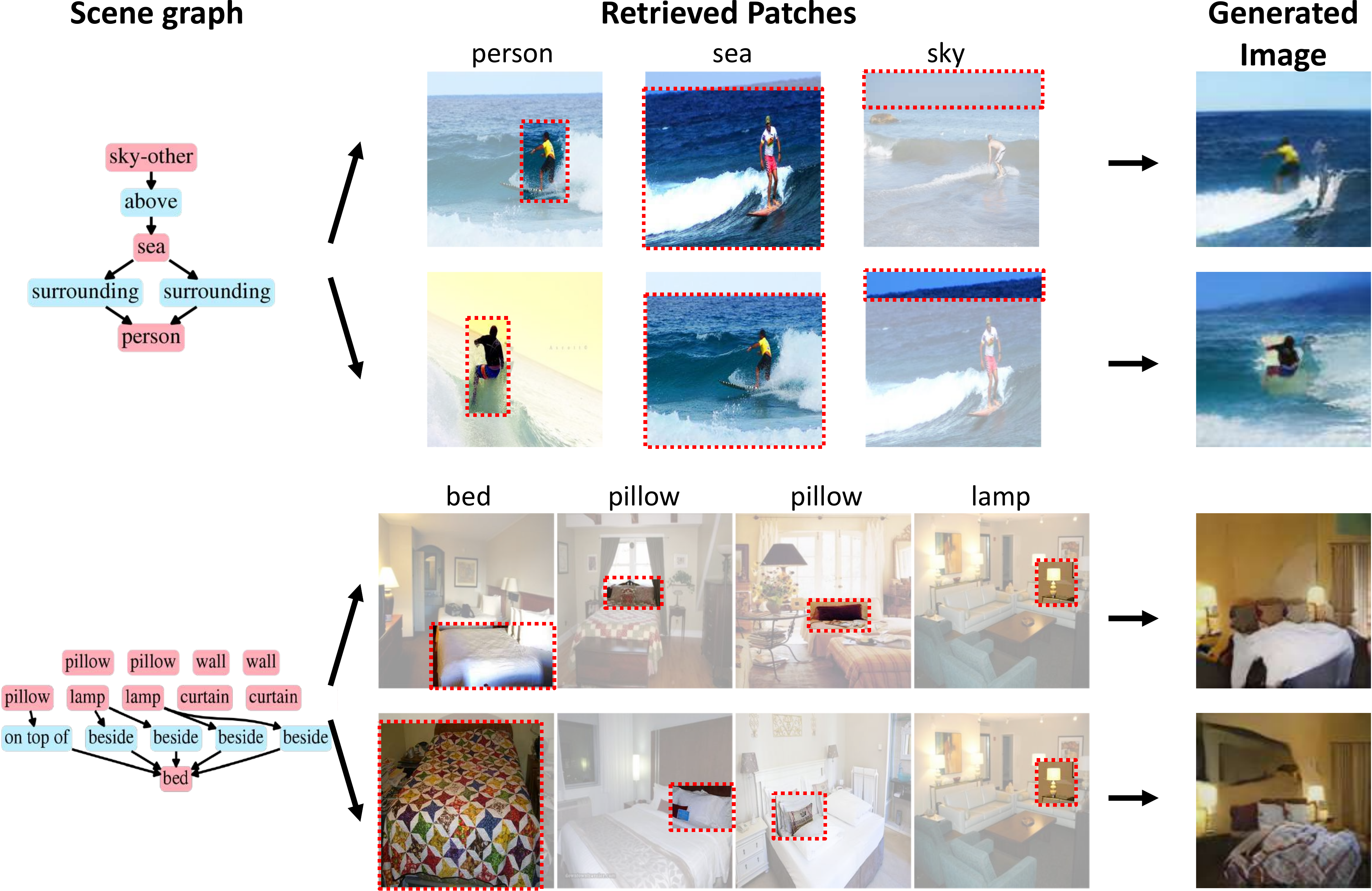}
    \vspace{-2.5mm}
    \caption{\textbf{Image synthesize from retrieved examples.} We propose the RetrieveGAN model that takes as input the scene graph description and learns to 1) select mutually compatible image patches via a differentiable retrieval process and 2) synthesize the output image from the retrieved patches.}
    \vspace{-9mm}
    \label{fig:teaser}
\end{figure}

\begin{abstract}
Image generation from scene description is a cornerstone technique for the controlled generation, which is beneficial to applications such as content creation and image editing.
In this work, we aim to synthesize images from scene description with retrieved patches as reference.
We propose a differentiable retrieval module.
With the differentiable retrieval module, we can (1) make the entire pipeline end-to-end trainable, enabling the learning of better feature embedding for retrieval; (2) encourage the selection of mutually compatible patches with additional objective functions.
We conduct extensive quantitative and qualitative experiments to demonstrate that the proposed method can generate realistic and diverse images, where the retrieved patches are reasonable and mutually compatible.

\end{abstract}

\section{Introduction}
\vspace{\secmargin}
\label{sec:intro}

Image generation from scene descriptions has received considerable attention.
Since the description often requests multiple objects in a scene with complicated relationships between objects, it remains challenging to synthesize images from scene descriptions.
The task requires not only the ability to generate realistic images but also the understanding of the mutual relationships among different objects in the same scene.
The usage of the scene description provides flexible user-control over the generation process and enables a wide range of applications in content creation~\cite{lee2019neural} and image editing~\cite{park2019SPADE}.

Taking advantage of generative adversarial networks (GANs)~\cite{goodfellow2014generative}, recent research employs conditional GAN for the image generation task.
Various conditional signals have been studied, such as scene graph~\cite{johnson2018image}, bounding box~\cite{zhao2019layout2im}, semantic segmentation map~\cite{park2019SPADE}, audio~\cite{lee2019dancing2music}, and text~\cite{xu2018attngan}.
A stream of work has been driven by parametric models that rely on the deep neural network to capture and model the appearance of objects~\cite{johnson2018image,xu2018attngan}.
Another stream of work has recently emerged to explore the semi-parametric model that leverages a memory bank to retrieve the objects for synthesizing the image~\cite{qi2018semi,yikang2019pastegan}.

In this work, we focus on the semi-parametric model in which a memory bank is provided for the retrieval purpose.
Despite the promising results, existing retrieval-based image synthesis methods face two issues.
First, the current models require pre-defined embeddings since the retrieval process is non-differentiable.
The pre-defined embeddings are independent of the generation process and thus cannot guarantee the retrieved objects are suitable for the surrogate generation task.
Second, oftentimes there are multiple objects to be retrieved given a scene description.
However, the conventional retrieval process selects each patch independently and thus neglect the subtle mutual relationship between objects.

We propose \emph{RetrieveGAN}, a conditional image generation framework with a differentiable retrieval process to address the issues.
First, we adopt the Gumbel-softmax~\cite{jang2016categorical} trick to make the retrieval process differentiable, thus enable optimizing the embedding through the end-to-end training.
Second, we design an iterative retrieval process to select a set of compatible patches~(\ie objects) for synthesizing a single image.
Specifically, the retrieval process operates iteratively to retrieve the image patch that is most compatible with the already selected patches.  
We propose a co-occurrence loss function to boost the mutual compatibility between the selected patches.
With the proposed differentiable retrieval design, the proposed RetrieveGAN is capable of retrieving image patches that 1) considers the surrogate image generation quality, and 2) are mutually compatible for synthesizing a single image.

We evaluate the proposed method through extensive experiments conducted on the COCO-stuff~\cite{caesar2018coco} and Visual Genome~\cite{krishna2017visual} datasets.
We use three metrics, Fr\'echet Inception Distance (FID)~\cite{fid}, Inception Score (IS)~\cite{salimans2016is}, and the Learned Perceptual Image Patch Similarity (LPIPS)~\cite{zhang2018lpips}, to measure the realism and diversity of the generated images.
Moreover, we conduct the user study to validate the proposed method's effectiveness in selecting mutually compatible patches.

To summarize, we make the following contributions in this work:
\begin{itemize}[$\bullet$]
    \item We propose a novel semi-parametric model to synthesize images from the scene description. The proposed model takes advantage of the complementary strength of the parametric and non-parametric techniques.
    \item We demonstrate the usefulness of the proposed differentiable retrieval module. The differentiable retrieval process can be jointly trained with the image synthesis module to capture the relationships among the objects in an image.
    \item Extensive qualitative and quantitative experiments demonstrate the efficacy of the proposed method to generate realistic and diverse images where retrieved objects are mutually compatible.
\end{itemize}
\section{Related Work}
\vspace{\secmargin}
\label{sec:related}

\Paragraph{Conditional image synthesis.}
The goal of the generative models is to model a data distribution given a set of samples from that distribution. 
The data distribution is either modeled explicitly (\eg variational autoencoder~\cite{kingma2013auto}) or implicitly (\eg generative adversarial networks~\cite{goodfellow2014generative}).
On the basis of unconditional generative models, conditional generative models target synthesizing images according to additional context such as image~\cite{choi2020starganv2,DRIT_plus,MSGAN,tseng2020art,CycleGAN2017}, segmentation mask~\cite{huang2020semantic,park2019SPADE,wang2018pix2pixHD,zhu2020sean}, and text.
The text conditions are often expressed in two formats: natural language sentences~\cite{xu2018attngan,han2017stackgan} or scene graphs~\cite{johnson2018image}.
Particularly, the scene graph description is in a well-structured format (\ie a graph with a node representing objects and edges describing their relationship), which mitigates the ambiguity in natural language sentences.
In this work, we focus on using the scene graph description as our input for the conditional image synthesis.

\Paragraph{Image synthesis from scene descriptions.}
Most existing methods employ parametric generative models to tackle this task.
The appearance of objects and relationships among objects are captured via a graph convolution network~\cite{johnson2018image,li2019controllable} or a text embedding network~\cite{li2019object,text2scene2019,xu2018attngan,han2017stackgan,zhu2019dm}, then images are synthesized with the conditional generative approach.
However, current parametric models synthesize objects at pixel-level, thus failing to generate realistic images for complicated scene descriptions.
More recent frameworks~\cite{qi2018semi,yikang2019pastegan} adopt semi-parametric models to perform generation at patch-level based on reference object patches.
These schemes retrieve reference patches from an external bank and use them to synthesize the final images.
Although the retrieval module is a crucial component, existing works all use predefined retrieval modules that cannot be optimized during the training stage.
In contrast, we propose a novel semi-parametric model with a differentiable retrieval process that is end-to-end trainable with the conditional generative model.

\Paragraph{Image Retrieval.}
Image retrieval has been a classical vision problem with numerous applications such as product search~\cite{liu2016deepfashion,han2017automatic,ak2018learning}, multimodal image retrieval~\cite{vo2019composing,chen2020image}, image geolocalization~\cite{hays2008im2gps}, event detection~\cite{jiang2014easy}, among others.
Solutions based on deep metric learning use the triplet loss~\cite{hermans2017defense,vo2016localizing} or softmax cross-entropy objective~\cite{vo2019composing} to learn a joint embedding space between the query (\eg text, image, or audio) and the target images. 
However, there is no prior work studying learning retrieval models for the image synthesis task.
Different from the existing semi-parametric generative models~\cite{text2scene2019,yikang2019pastegan} that use the pre-defined (or fixed) embedding to retrieve image patches, we propose a differentiable retrieval process that can be jointly optimized with the conditional generative model.
\begin{figure}[t]
    \centering
    \subfloat[RetrieveGAN]{
    \includegraphics[width=0.99\linewidth]{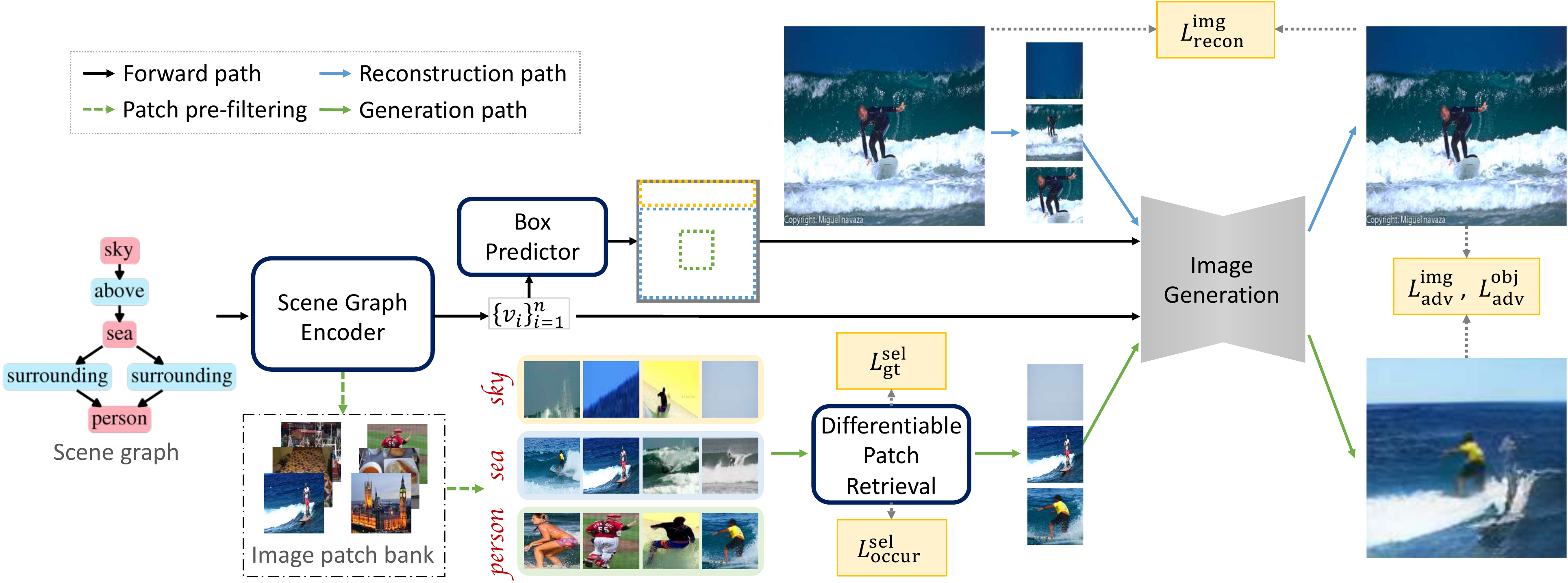}
    }\\
    \subfloat[Iterative differentiable patch retrieval]{
    \includegraphics[width=0.85\linewidth]{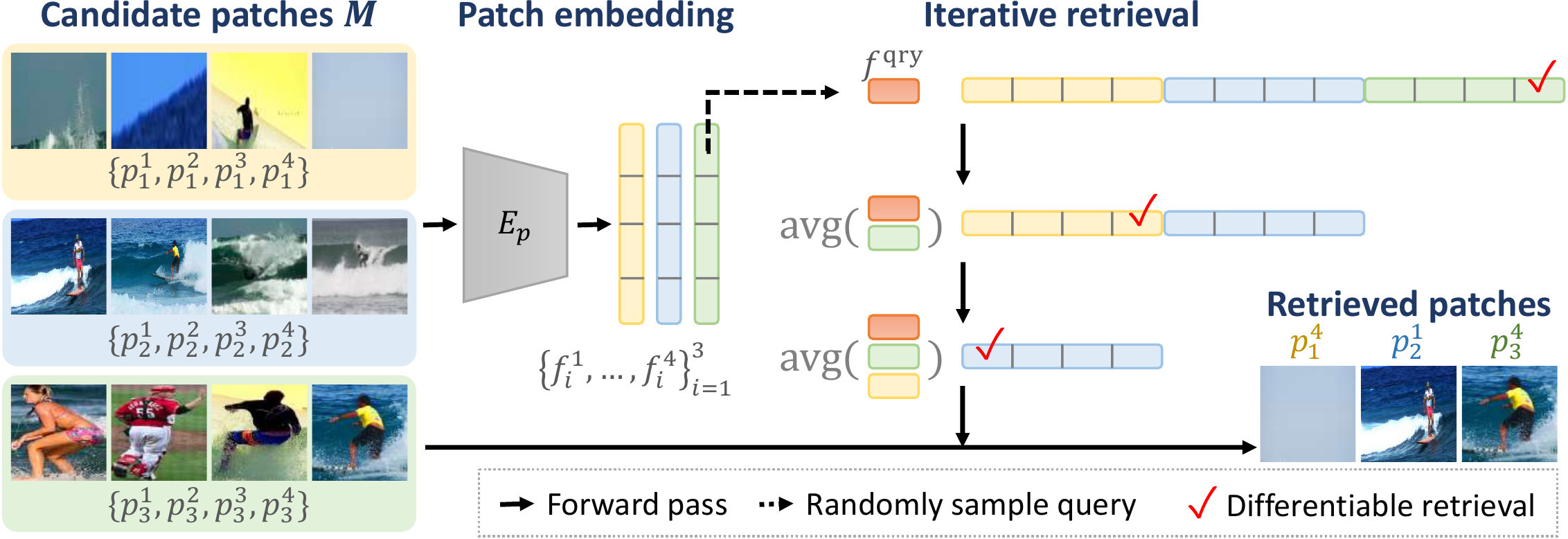}
    }
    \vspace{-2mm}
    \caption{\textbf{Method overview.} (a) Our model takes as input the scene graph description and sequentially performs scene graph encoding, patch retrieval, and image generation to synthesize the desired image. 
    (b) Given a set of candidate patches, we first extract the corresponding patch features using the patch embedding function. 
    We then randomly select a patch feature as the query feature for the iterative retrieval process. 
    At each step of the iterative procedure, we select the patch that is most compatible with the already selected patches. 
    The iteration ends as all the objects are assigned with a selected patch.}
    \label{fig:method}
    \vspace{\figmargin}
\end{figure}
\section{Methodology}
\vspace{\secmargin}
\label{sec:method}

\subsection{Preliminaries}
Our goal is to synthesize an image $x\in \mathbb{R}^{H\times{W}\times{3}}$ from the input scene graph $g$ by compositing appropriate image patches retrieved from the image patch bank.
As the overview shown in~\figref{method}, the proposed RetrieveGAN framework consists of three stages: scene graph encoding, patch retrieval, and image generation.
The scene graph encoding module processes the input scene graph $g$, extracts features, and predicts bounding box coordinates for each object $o_i$ defined in the scene graph.
The patch retrieval module then retrieves an image patch for each object $o_i$ from the image patch bank.
The goal of the retrieval module is to maximize the compatibility of all retrieved patches, thus improving the quality of the image synthesized by the subsequent image generation module.
Finally, the image generation module takes as input the selected patches along with the predicted bounding boxes to synthesize the final image.

\Paragraph{Scene graph.} Serving as the input data to our framework, the scene graph representation~\cite{johnson2015image} describes the objects in a scene and the relationships between these objects.
We denote a set of object categories as $\mathcal{C}$  and relation categories as $\mathcal{R}$.
A scene graph $g$ is then defined as a tuple $(\{o_i\}^n_{i=1}, \{e_i\}^m_{i=1})$, where $\{o_i | o_i \in \mathcal{C} \}^n_{i=1}$ is a set of objects in the scene.
The notation $\{e_i\}^m_{i=1}$ denotes a set of direct edges in the form of $e_i = (o_j, r_k, o_t)$ where $o_j, o_t \in \mathcal{C}$ and $r_k \in \mathcal{R}$. 

\Paragraph{Image patch bank.} The second input to our model is the memory bank consisting of all available real image patches for synthesizing the output image.
Following PasteGAN~\cite{yikang2019pastegan}, we use the ground-truth bounding box to extract the images patches $M = \{p_i \in \mathbb{R}^{h\times{w}\times{3}}\}$ from the training set.
Note that we relax the assumption in PasteGAN and do not use the ground-truth mask to segment the image patches in the COCO-Stuff~\cite{caesar2018coco} dataset.

\subsection{Scene Graph Encoding}
The scene graph encoding module aims to process the input scene graph and provides necessary information for the later patch retrieval and image generation stages.
We detail the process of scene graph encoding as follows:

\Paragraph{Scene graph encoder.} Given an input scene graph $g=(\{o_i\}^n_{i=1}, \{e_i\}^m_{i=1})$, the scene graph encoder $E_\mathrm{g}$ extracts the object features, namely $\{v_i\}^n_{i=1} = E((\{o_i\}^n_{i=1}, \{e_i\}^m_{i=1}))$. 
%
%
Adopting the strategy in sg2im~\cite{johnson2018image}, we construct the scene graph encoder with a series of graph convolutional networks~(GCNs).
We further discuss the detail of the scene graph encoder in the supplementary document.

\Paragraph{Bounding box predictor.}
For each object $o_i$, the bounding box predictor learns to predict the bounding box coordinates $\hat{b}_i=(x_0, y_0, x_1, y_1)$ from the object features $v_i$.
We use a series of fully-connected layers to build the predictor.

\Paragraph{Patch pre-filtering.} 
Since there are a large number of image patches in the image patch bank, performing the retrieval on the entire bank online is intractable in practice due to the memory limitation.
We address this problem by pre-filtering a set of $k$ candidate patches $M(o_i) = \{p^1_i, p^2_i,\cdots, p^k_i\}$ for each object $o_i$.
And the later patch retrieval process is conducted on the pre-filtered candidate patches as opposed to the entire patch bank. 
To be more specific, we use the pre-trained GCN in sg2im~\cite{johnson2018image} to obtain the candidate patches for each object. We use the corresponding scene graph to compute the GCN feature.
The computed GCN feature is used to select similar candidate patches $M(o_i)$ with respect to the negative $\ell_2$ distance.

\subsection{Patch Retrieval}
The patch retrieval aims to select a number of mutually compatible patches for synthesizing the output image. 
We illustrate the overall process on the bottom side of \figref{method}.
Given the pre-filtered candidate patches $\{M(o_i)\}^n_{i=1}$, we first use a patch embedding function $E_p$ to extract the patch features.
Starting with a randomly sampled patch feature as a query, we propose an iterative retrieval process to select compatible patches for all objects.
In the following, we 1) describe how a single retrieval is operated, 2) introduce the proposed iterative retrieval process, and 3) discuss the objective functions used to facilitate the training of the patch retrieval module.

\Paragraph{Differentiable retrieval for a single object.}
Given the query feature $f^\mathrm{qry}$, we aim to sample a single patch from the candidate set $M(o)=\{p^1,p^2,\cdots,p^k\}$ for object $o$. 
Let $\pi \in \mathbb{R}_{>0}^{k}$ be the categorical variable with probabilities $P(x=i) \propto \pi_i$ which indicates the probability of selecting the $i$-th patch from the bank. 
To compute $\pi_i$, we calculate the $\ell_2$ distance between the query feature and the corresponding patch feature, namely $\pi_i = e^{-\lVert f_\mathrm{qry} - E_p(p^i;\theta_{E_p}) \rVert_2}$, where $E_p$ is the embedding function and $\theta_{E_p}$ is the learnable mode parameter.
The intuition is that the candidate patch with smaller feature distance to the query feature should be sampled with higher probability. 
By optimizing $\theta_{E_p}$ with our loss functions, we hope our model is capable of retrieving compatible patches.
As we are sampling from a categorical distribution, we use the Gumbel-Max trick \cite{jang2016categorical} to sample a single patch:
\begin{align}
\arg\max_i [P(x=i)]= \arg\max_i [g_i + \log \pi_i]=\arg\max_i[\hat{\pi}_i],
\label{eq:gumbel}
\end{align}
where $g_i = -\log(-\log(u_i))$ is the re-parameterization term and $u_i \sim \text{Uniform}(0,1)$.
To make the above process differentiable, the argmax operation is approximated with the continuous softmax operation:
\begin{align}
s = \text{softmax}(\hat{\pi}) = \frac{\exp(\hat{\pi}_i/\tau)}{{\sum_{q=1}^k}\exp(\hat{\pi}_q/\tau)},
\label{eq:softmax}
\end{align}
where $\tau$ is the temperature controlling the degree of the approximation.\footnote{When $\tau$ is small, we found it is useful to make the selection variable $s$ uni-modal. This can also be achieved by post-processing (\eg thresholding) the softmax outputs.}

\Paragraph{Iterative differentiable retrieval for multiple objects.}
Rather than retrieving only a single image patch, the proposed framework needs to select a subset of $n$ patches for the $n$ objects defined in the input scene graph.
Therefore, we adopt the weighted reservoir sampling strategy~\cite{xie2019reparameterizable} to perform the subset sampling from the candidate patch sets. Let $M = \{p_i | i = 1, \ldots, n \times k\}$ denote a multiset (with possible duplicated elements) consisting of all candidates patches in which $n$ is the number of objects and $k$ is the size of each candidate patch set.
We leave the preliminaries on weighted reservoir sampling in the supplementary materials.
In our problem, we first compute the vector $\hat{\pi}_i$ defined in~\eqnref{gumbel} for all patches.
We then iteratively apply $n$ softmax operations over $\hat{\pi}$ to approximate the top-$k$ selection.
Let $\hat{\pi}_i^{(j)}$ denote the probability of sampling patch $p_i$ at iteration $j$ and $\hat{\pi}_i^{(1)} \leftarrow \hat{\pi}_i$. The probability is iteratively updated by:
\begin{align}
\label{eq:iter_selection}
\hat{\pi}_i^{(j+1)} \leftarrow \hat{\pi}_i^{(j)} + \log(1-s_i^{(j)}),
\end{align}
where $s_i^{(j)}=\text{softmax}(\hat{\pi}^{(j)})_i$ computed by \eqref{eq:softmax}. 
Essentially, \eqref{eq:iter_selection} sets the probability of the selected patch to negative infinity, thus ensures this patch will not be chosen again.
After $n$ iterations, we compute the relaxed $n$-hot vector $s = \sum_{j=1}^{n} s^{(j)}$, where $s_i \in [0,1]$ indicates the score of selecting the $i$-th patch and $\sum_{i=1}^{|M|} s_i= n$.
The entire process is differentiable with respect to the model parameters.

We make two modifications to the above iterative process based on practical consideration.
First, our candidate multiset $M = \{p_i\}_{i=1}^{n \times k}$ is formed by $n$ groups of pre-filtered patches where every object has a group $k$ patches. Since we are only allowed to retrieve a single patch from a group, we modify~\eqnref{iter_selection} by:
\begin{align}
\label{eq:iter_group_selection}
\hat{\pi}_i^{(j+1)} \leftarrow \hat{\pi}_i^{(j)} + \log(1-\max_{t}[s_t^{(j)}]) \quad \forall t \text{  such that  }  m^{-1}(p_i) = t,
\end{align}
where we denote $m^{-1}(p_j)=i$ if patch $p_j$ in $M$ is pre-fetched by the object $o_i$. \eqref{eq:iter_group_selection} uses max pooling to disable selecting multiple patches from the same group.
Second, to incorporate the prior knowledge that compatible images patches tend to lie closer in the embedding space, we use a greedy strategy to encourage selecting image patches that are compatible with the already selected ones. We detail this process in~\figref{method}(b).
To be more specific, at each iteration, the features of the selected patches are aggregated by average pooling to update the query $f^\mathrm{qry}$. $\pi$ and $\hat{\pi}$ is also recomputed accordingly after the query update.
This leads to a greedy strategy encouraging the selected patches to be visually or semantically similar in the feature space. 
We summarize the overall retrieval process in Algorithm~\ref{alg:iterative}.
\begin{algorithm}[t]
\SetKwData{Left}{left}\SetKwData{This}{this}\SetKwData{Up}{up}
\SetKwInOut{Input}{Input}\SetKwInOut{Output}{Output}
\LinesNumbered
\DontPrintSemicolon
\small
\Input{Candidate patches $M = \{p_i\}_{i=1}^{n \times k}$ for $n$ objects and each object has $k$ pre-filtered patches.}
\Output{relaxed $n$-hot vector $s$ where $\sum_{i=1}^{|M|} s_i = n$ and $0 \le s_i \le 1$.}
\BlankLine
\lFor{$i = 1, \ldots, |M|$ }{
    $f_i=E_p(p_i)$ \tcp*[h]{Get patch features}
}
\BlankLine
Randomly select a patch feature to initialize the query $f^\mathrm{qry}$\;
\BlankLine
\tcp{Iterative patch retrieval}
\For{$t = 1, \ldots, n$}{
    \For{$i = 1, \ldots, |M|$}{
        \BlankLine
        $\pi_i=e^{-\lVert f_i - f^\mathrm{qry}\rVert_2}$ \tcp{Calculate $\pi$ according to the query} 
        
        \BlankLine
        \tcp{Gumbel-Max trick}
        $u_i \leftarrow \text{Uniform}(0,1)$ \;
        $\hat{\pi}_i \leftarrow -\log(-\log(u_i)) + \log(\pi_i)$ \;
        
        \BlankLine
        \tcp{Disable other patches in the selected group}
        $\hat{\pi}_i \leftarrow \hat{\pi}_i + \log(1-\max_{j}[s_j^{(t-1)}]) \quad \forall j \text{  such that  }  m^{-1}(p_i) = j$\;
    }
    \BlankLine
    \lFor{$i = 1, \ldots, |M|$}{
        $s_i^{(t)} = \frac{\exp(\hat{\pi}_i/\tau)}{{\sum_{q=1}^{|M|}}\exp(\hat{\pi}_q/\tau)}$\tcp*[h]{Softmax operation}
    }
    \BlankLine
    $f^\mathrm{qry}=\mathrm{avg}(f^\mathrm{qry},\sum_{i=1}^{|M|}s_i^{(t)}f_i)$ \tcp{Update the query}
}
\Return the relaxed $n$-hot vector $s^{(n)}$
\caption{Iterative Differential Retrieval}
\label{alg:iterative}
\end{algorithm}

As the retrieval process is differentiable, we can optimize the retrieval module~(\ie patch embedding function $E_p$) with the loss functions~(\eg adversarial loss) applied to the following image generation module.
Moreover, we incorporate two additional objectives to facilitate the training of iterative retrieval process: ground-truth selection loss $L^\mathrm{sel}_\mathrm{gt}$ and co-occurrence loss $L^\mathrm{sel}_\mathrm{occur}$.

\Paragraph{Ground-truth selection loss.}
As the ground-truth patches are available at the training stage, we add them to the candidate set $M$.
Given one of the ground-truth patch features as the query feature $f^\mathrm{qry}$, the ground-truth selection loss $L^\mathrm{sel}_\mathrm{gt}$ encourages the retrieval process to select the ground-truth patches for the other objects in the input scene graph.

\Paragraph{Co-occurrence penalty.}
We design a co-occurrence loss to ensure the mutual compatibility between the retrieved patches.
The core idea is to minimize the distances between the retrieved patches in a co-occurrence embedding space.
Specifically, we first train a co-occurrence embedding function $F_\mathrm{occur}$ using the patches cropped from the training images with the triplet loss~\cite{wang2017transitive}.
The distance on the co-occurrence embedding space between the patches sampled from the same image is minimized, while the distance between the patches cropped from the different images is maximized.
Then the proposed co-occurrence loss is the pairwise distance between the retrieved patches on the co-occurrence embedding space:
\begin{align}
L^\mathrm{sel}_\mathrm{occur}=\sum_{i,j} d(F_\mathrm{occur}(p_i),F_\mathrm{occur}(p_j)),
\end{align}
where $p_i$ and $p_j$ are the patches retrieved by the iterative retrieval process.

\Paragraph{Limitations~\vs~advantages.}
The size of the candidate patches considered by the proposed retrieval process is currently limited by the GPU memory.
Therefore, we cannot perform the differentiable retrieval over the entire memory bank.
Nonetheless, the differentiable mechanism and iterative design enable us to train the retrieval process using the abovementioned loss functions that maximize the mutual compatibility of the selected patches.

\subsection{Image Generation}
Given selected patches after the differentiable patch retrieval process, the image generation module synthesizes the realistic image with the selected patches as reference.
We adopt a similar architecture to PasteGAN~\cite{yikang2019pastegan} as our image generation module.
Please refer to the supplementary materials for details regarding the image generation module.
We use two discriminators $D_\mathrm{img}$ and $D_\mathrm{obj}$ to encourage the realism of the generated images on the image-level and object-level, respectively.
Specifically, the adversarial loss can be expressed as:
\begin{equation}
\begin{aligned}
& L^\mathrm{img}_\mathrm{adv}=\mathbb{E}_{x}[\log{D_\mathrm{img}(x)}]+\mathbb{E}_{\hat{x}}[\log{(1-D_\mathrm{img}(\hat{x}))}], \\
& L^\mathrm{obj}_\mathrm{adv} = \mathbb{E}_{p}[\log{D_\mathrm{obj}(p)}]+\mathbb{E}_{\hat{p}}[\log{(1-D_\mathrm{obj}(\hat{p}))}],
\end{aligned}
\end{equation}
where $x$ and $p$ denote the real image and patch, whereas $\hat{x}$ and $\hat{p}$ represent the generated image and the patch crop from the generated image, respectively.

\subsection{Training objective functions}
In addition to the abovementioned loss functions, we use the following loss functions during the training phase:

\Paragraph{Bounding box regression loss.} 
We penalize the prediction of the bounding box coordinates with $\ell1$ distance $L_\mathrm{bbx}=\sum_{i=1}^{n}\lVert b_i - \hat{b_i} \rVert_{1}$.

\Paragraph{Image reconstruction loss.}
Given the ground-truth patches and the ground-truth bounding box coordinates, the image generation module should recover the ground-truth image.
The loss $L^\mathrm{img}_\mathrm{recon}$ is an $\ell1$ distance measuring the difference between the recovered and ground-truth images.

\Paragraph{Auxiliary classification loss.}
We adopt the auxiliary classification loss $L^\mathrm{obj}_\mathrm{ac}$ to encourage the generated patches to be correctly classified by the object discriminator $D_{obj}$.

\Paragraph{Perceptual loss.}
The perceptual loss is computed as the distance in the pre-trained VGG~\cite{simonyan2014very} feature space.
We apply the perceptual losses $L^\mathrm{img}_\mathrm{p},L^\mathrm{obj}_\mathrm{p}$ on both image and object levels to stabilize the training procedure.

The full loss functions for training our model is:
\begin{equation}
\begin{aligned}
L=&\lambda^\mathrm{sel}_\mathrm{gt}L^\mathrm{sel}_\mathrm{gt} +  \lambda^\mathrm{sel}_\mathrm{occur}L^\mathrm{sel}_\mathrm{occur} + \lambda^\mathrm{img}_\mathrm{adv}L^\mathrm{img}_\mathrm{adv} + \lambda^\mathrm{img}_\mathrm{recon}L^\mathrm{img}_\mathrm{recon} +
\lambda^\mathrm{img}_\mathrm{p}L^\mathrm{img}_\mathrm{p} + \\
&\lambda^\mathrm{obj}_\mathrm{adv}L^\mathrm{obj}_\mathrm{adv} + \lambda^\mathrm{obj}_\mathrm{ac}L^\mathrm{obj}_\mathrm{ac} +  \lambda^\mathrm{obj}_\mathrm{p}L^\mathrm{obj}_\mathrm{p} +\lambda_\mathrm{bbx}L_\mathrm{bbx},
\end{aligned}
\end{equation}
where $\lambda$ controls the importance of each loss term.
We describe the implementation detail of the proposed approach in the supplementary document.
\section{Experimental Results}
\vspace{\secmargin}
\label{sec:exp}

\Paragraph{Datasets.}
The COCO-Stuff~\cite{caesar2018coco} and Visual Genome~\cite{krishna2017visual} datasets are standard benchmark datasets for evaluating scene generation models~\cite{johnson2018image,yikang2019pastegan,zhao2019layout2im}.
We use the image resolution of $128\times 128$ for all the experiments.
Except for the image resolution, we follow the protocol in sg2im~\cite{johnson2018image} to pre-process and split the dataset.
Different from the PasteGAN~\cite{yikang2019pastegan} approach, we do not access the ground-truth mask for segmenting the image patches.

\begin{table*}[t]
    \caption{\textbf{Quantitative comparisons.}
    We evaluate all methods on the COCO-Stuff and Visual Genome datasets using the FID, IS, and DS metrics.
    The first row shows the results of models that predict bounding boxes during the inference time.
    The second row shows the results of models that take ground-truth bounding as inputs during the inference time.
    }
    \label{tab:Quan}
    \centering
    \begin{tabular}{l ccc ccc } 
	    \toprule
		Datasets & \multicolumn{3}{c}{COCO-Stuff}&\multicolumn{3}{c}{Visual Genome}
		\\  \cmidrule(lr){2-4} \cmidrule(lr){5-7} 
		& FID $\downarrow$ & IS $\uparrow$ & DS $\uparrow$ 	& FID $\downarrow$ & IS $\uparrow$ & DS $\uparrow$ \\
		
		sg2im~\cite{johnson2018image} &
		$136.8$ & $4.1${\tiny$\pm0.1$} & $0.02${\tiny$\pm0.0$} & $126.9$ & $5.1${\tiny$\pm0.1$} & $0.11${\tiny$\pm0.1$}\\
		AttnGAN~\cite{xu2018attngan} &
		$72.8$ & $8.4${\tiny$\pm0.2$} & $0.14${\tiny$\pm0.1$} & $114.6$ & $\textbf{10.4}${\tiny$\pm0.2$} & $\underline{0.27}${\tiny$\pm0.2$} \\
		PasteGAN~\cite{yikang2019pastegan} &
		$\underline{59.8}$ & $\underline{8.8}${\tiny$\pm0.3$} & $\textbf{0.43}${\tiny$\pm0.1$} & $\underline{81.8}$ & $6.7${\tiny$\pm0.2$} & $\textbf{0.30}${\tiny$\pm0.1$} \\
		RetrieveGAN (Ours) &
		$\textbf{43.2}$ & $\textbf{10.6}${\tiny$\pm0.6$} & $\underline{0.34}${\tiny$\pm0.1$} & $\textbf{70.3}$ & $\underline{7.7}${\tiny$\pm0.1$} & $0.24${\tiny$\pm0.1$} \\
		\midrule
		sg2im (GT) &
		$79.9$ & $8.5${\tiny$\pm0.1$} & $0.02${\tiny$\pm0.0$} & $111.9$ & $5.8${\tiny$\pm0.1$} & $0.13${\tiny$\pm0.1$}\\
		layout2im~\cite{zhao2019layout2im} &
		$\underline{45.3}$ & $\underline{10.2}${\tiny$\pm0.6$} & $\underline{0.29}${\tiny$\pm0.1$} & $\textbf{44.0}$ & $\textbf{9.3}${\tiny$\pm0.4$} & $\textbf{0.29}${\tiny$\pm0.1$}\\
		PasteGAN (GT) &
		$54.9$ & $9.6${\tiny$\pm0.2$} & $\textbf{0.38}${\tiny$\pm0.1$} & $68.1$ & $6.7${\tiny$\pm0.1$} & $\underline{0.28}${\tiny$\pm0.1$}\\
		RetrieveGAN (GT) &
		$\textbf{42.7}$ & $\textbf{10.7}${\tiny$\pm0.1$} & $0.21${\tiny$\pm0.1$} & $\underline{46.3}$ & $\underline{9.1}${\tiny$\pm0.1$} & $0.23${\tiny$\pm0.1$} \\
        \midrule
        Real data & 
        $6.8$ & $24.3${\tiny$\pm0.3$} & - & $6.9$ & $24.1${\tiny$\pm0.4$} & - \\
		\bottomrule
    \end{tabular}

\end{table*}

\Paragraph{Evaluated methods.} We compare the proposed approach to three parametric generation models and one semi-parametric model in the experiments:
\begin{compactitem}
\item \tb{sg2im}~\cite{johnson2018image}: 
The sg2im framework takes as input a scene graph and learns to synthesize the corresponding image.
%
\item \tb{AttnGAN}~\cite{xu2018attngan}: As the AttnGAN method synthesizes the images from text, we convert the scene graph to the corresponding text description. Specifically, we convert each relationship in the graph into a sentence, and link every sentence via the conjunction word ``and". We train the AttnGAN model on these converted sentences.

\item \tb{layout2im}~\cite{zhao2019layout2im}: The layout2im scheme takes as input the ground-truth bounding boxes to perform the generation.
For a fair comparison, we use the ground-truth bounding box coordinate as the input data for other methods, which we denote GT in the experimental results.

\item \tb{PasteGAN}~\cite{yikang2019pastegan}:
The PasteGAN approach is most related to our work as it uses the pre-trained embedding function to retrieve candidate patches.
%
%


\end{compactitem}

\begin{table*}[t]
    \centering
    \caption{\textbf{Ablation studies.}
    We conduct ablation studies on two loss functions added upon the proposed retrieval module.
    }
        \label{tab:ablation}
    \begin{tabular}{l cc ccc } 
	    \toprule
		Datasets & & & \multicolumn{3}{c}{COCO-Stuff} \\
		\cmidrule(lr){4-6}
		 & $\mathcal{L}^\mathrm{sel}_\mathrm{gt}$ & $\mathcal{L}^\mathrm{sel}_\mathrm{occur}$ & FID $\downarrow$ & IS $\uparrow$ & DS $\uparrow$ 	 \\
		RetrieveGAN & - & - &
		$56.8$ & $8.8${\tiny$\pm0.3$} & $0.30${\tiny$\pm0.1$} \\
		RetrieveGAN & \checkmark & - &
		$47.8$ & $9.7${\tiny$\pm0.2$} & $\textbf{0.36}${\tiny$\pm0.1$} \\
		RetrieveGAN & - & \checkmark &
		$52.8$ & $9.8${\tiny$\pm0.2$} & $0.29${\tiny$\pm0.1$} \\
		RetrieveGAN & \checkmark  & \checkmark &
		$\textbf{43.2}$ & $\textbf{10.6}${\tiny$\pm0.6$} & $0.34${\tiny$\pm0.1$} \\
        \midrule
        Real data &  &  &
        $6.8$ & $24.3${\tiny$\pm0.3$}  & - \\
		\bottomrule
    \end{tabular}
    \vspace{\tabmargin}
\end{table*}

\Paragraph{Evaluation Metrics.} We use the following metrics to measure the realism and diversity of the generated images:
\begin{compactitem}
\item \tb{Inception Score (IS).}
Inception Score~\cite{salimans2016is} uses the Inception V3~\cite{szegedy2016rethinking} model to measure the visual quality of the generated images.
\item \tb{Fr\'echet Inception Distance (FID).}
Fr\'echet Inception Distance~\cite{fid} measures the visual quality and diversity of the synthesized images. 
We use the Inception V3 model as the feature extractor.
\item \tb{Diversity (DS).}
We use the AlexNet model to explicitly evaluate the diversity by measuring the distances between the features of the images using the Learned Perceptual Image Patch Similarity (LPIPS)~\cite{zhang2018lpips} metric.
\end{compactitem}

\begin{figure}[t]
    \centering
    \includegraphics[width=0.7\linewidth]{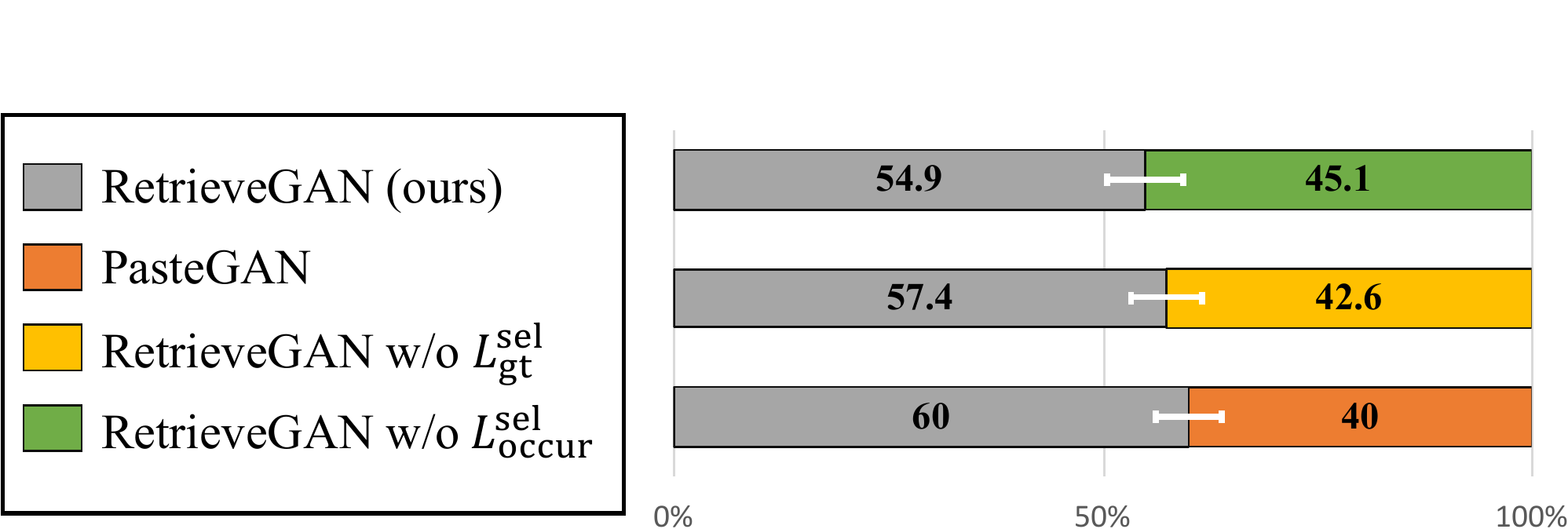}
     \caption{\textbf{User study.}
     We conduct the user study to evaluate the mutual compatibility of the selected patches.
     }
    \label{fig:userstudy}
    \vspace{\figmargin}
\end{figure}
\begin{figure}[t]
    \centering
    \includegraphics[width=0.99\linewidth]{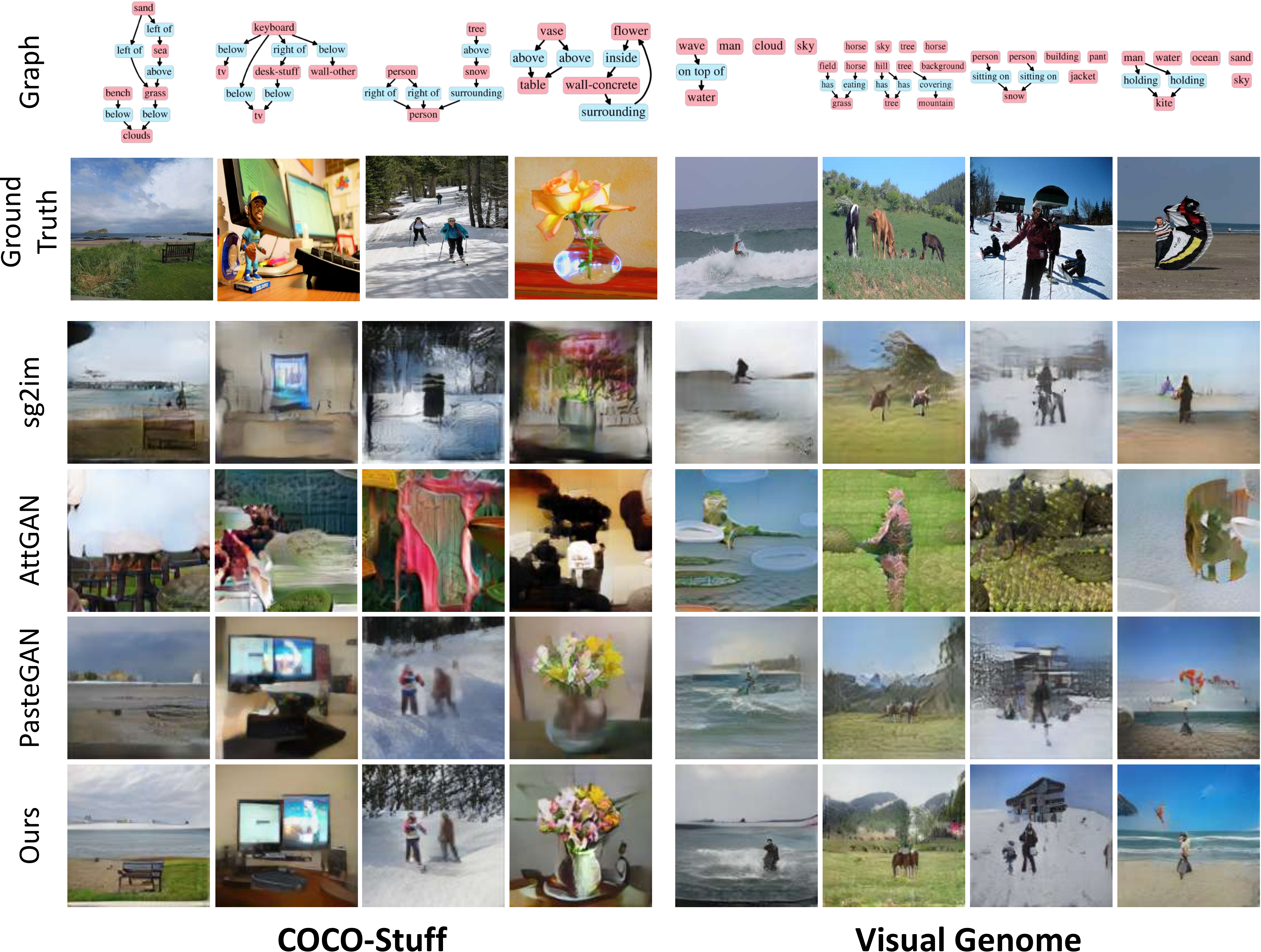}
    \caption{\textbf{Sample generation results.} We show example results on the COCO-Stuff~(\textit{left}) and Visual Genome~(\textit{right}) datasets.
    The object locations in each image are predicted by models.
    }
    \label{fig:result1}
    \vspace{\figmargin}
\end{figure}

\begin{figure}[t]
    \centering
    \includegraphics[width=0.99\linewidth]{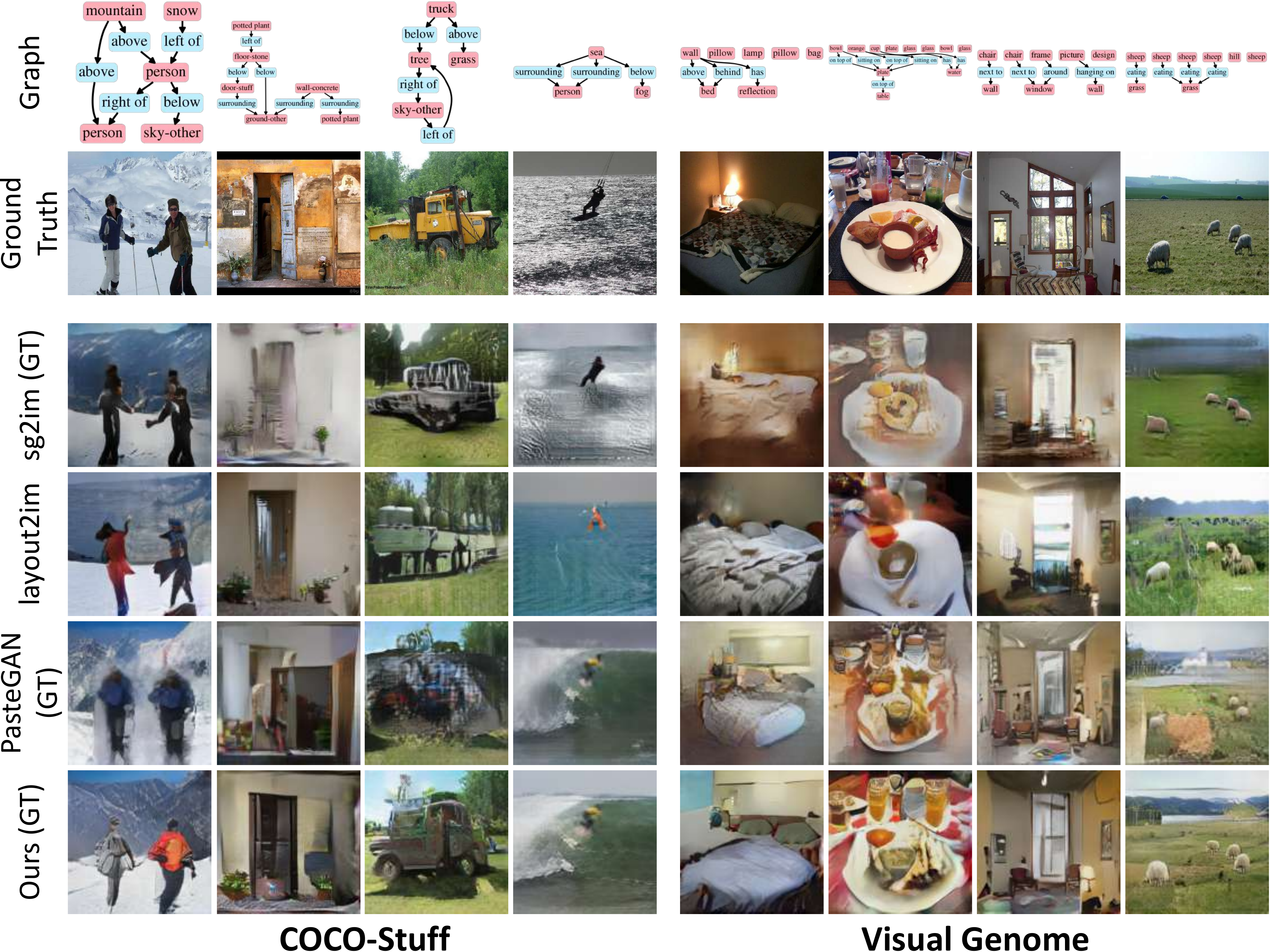}
    \caption{\textbf{Sample generation results.} We show example results on the COCO-Stuff~(\textit{left}) and Visual Genome~(\textit{right}) datasets. 
    The object locations in each image are given as additional inputs.
    }
    \label{fig:result2}
    \vspace{\figmargin}
\end{figure}

\subsection{Quantitative Evaluation}
\label{sec:exp_quant}
\Paragraph{Realism and diversity.}
We evaluate the realism and diversity of all methods using the IS, FID, and DS metrics.
To have a fair comparison with different methods, we conduct the evaluation using two different settings.
First, bounding boxes of objects are predicted by models.
Second, ground-truth bounding boxes are given as inputs in addition to the scene graph.
The results of these two settings are shown in the first and second row of \tabref{Quan}, respectively.
Since the patch retrieval process is optimized to consider the generation quality during the training stage, our approach performs favorably against the other algorithms in terms of realism.
On the other hand, as we can sample different query features for the proposed retrieval process, our model synthesizes comparably diverse images compared to the other schemes.

\begin{figure}[th]
    \centering
    \includegraphics[width=0.99\linewidth]{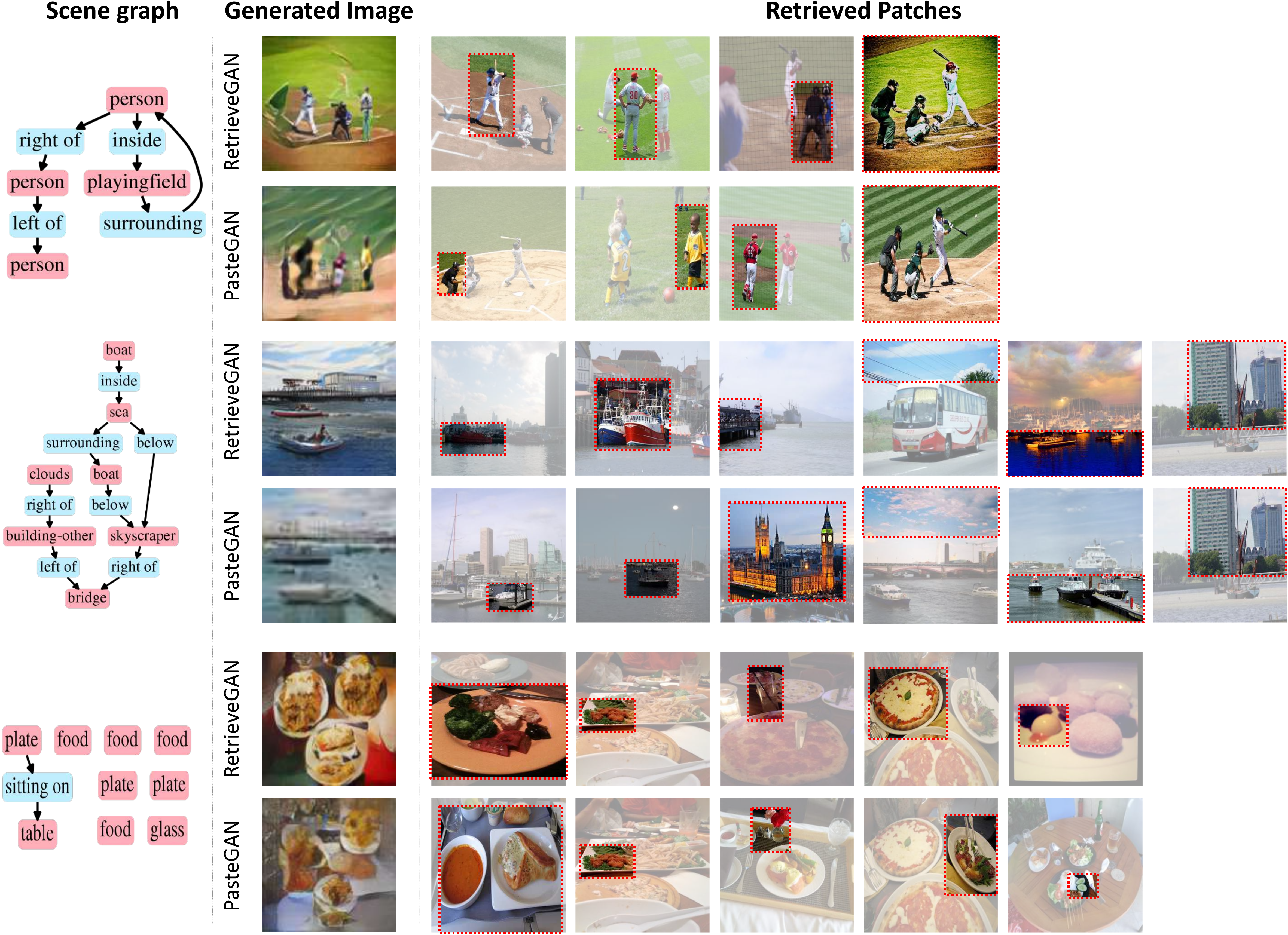}
    \vspace{-1.5mm}
    \caption{\textbf{Retrieved patches.}
    For each sample, we show the retrieved patches which are used to guide the following image generation process.
    We also show the original image of each selected patch for more clear visualization.
    }
    \label{fig:retrieved}
\end{figure}

Moreover, there are two noteworthy observations.
First, the proposed RetrieveGAN has similar performance in both settings on the COCO-Stuff dataset, but has significant improvement using ground-truth bounding boxes on the Visual Genome dataset.
The reason for the inferior performance on the Visual Genome dataset without using ground-truth bounding boxes is due to the existence of lots of isolated objects (\ie objects that have no relationships to other objects) in the scene graph annotation~(\eg the last scene graph in~\figref{retrieved}), which greatly increase the difficult of predicting reasonable bounding boxes.
Second, on the Visual Genome dataset, AttnGAN outperforms the proposed method on the IS and DS metrics, while performs significantly worse than the proposed method on the FID metric.
Compared to the FID metric, the IS score has the limitation that it is less sensitive to the mode collapse problem.
The DS metric only measures the feature distance without considering visual quality.
The results from AttnGAN shown in \figref{result1} also support our observation.

\Paragraph{Patch compatibility.}
The proposed differentiable retrieval process aims to improve the mutual compatibility among the selected patches.
We conduct a user study to evaluate the patch compatibility.
For each scene graph, we present two sets of patches selected by different methods, and ask users ``which set of patches are more mutually compatible and more likely to coexist in the same image?''.
\figref{userstudy} presents the results of the user study.
The proposed method outperforms PasteGAN, which uses a pre-defined patch embedding function for retrieval.
The results also validate the benefits of the proposed ground-truth selection loss and co-occurrence loss.

\Paragraph{Ablation study.}
We conduct an ablation study on the COCO-Stuff dataset to understand the impact of each component in the proposed design.
The results are shown in~\tabref{ablation}.
As the ground-truth selection loss and the co-occurrence penalty maximize the mutual compatibility of the selected patches, they both improve the visual quality of the generated images.

\subsection{Qualitative Evaluation}

\Paragraph{Image generation.}
We qualitatively compare the visual results generated by different methods.
We show the results on the COCO-Stuff (left column) and the Visual Genome (right column) datasets under two settings of using predicted (\figref{result1}) and ground-truth (\figref{result2}) bounding boxes.
The sg2im and layout2im methods can roughly capture the appearance of objects and mutual relationships among objects. 
However, the quality of generated images in complicated scenes is limited.
Similarly, the AttnGAN model cannot handle scenes with complex relationships well.
The overall image quality generated by the PasteGAN scheme is similar to that by the proposed approach, yet the quality is affected by the compatibility of the selected patches~(\eg the third result on COCO-Stuff in \figref{result2}).

\Paragraph{Patch retrieval.}
To better visualize the source of retrieved patches, we present the generated images as well as the original images of selected patches in \figref{retrieved}.
The proposed method can tackle complex scenes where multiple objects are present.
With the help of selected patches, each object in the generated images has a clear and reasonable appearance (\eg the boat in the second row and the food in the third row).
Most importantly, the retrieved patches are mutually compatible, thanks to the proposed iterative and differentiable retrieval process.
As shown in the first example in \figref{retrieved}, the selected patches are all related to baseball, while the PasteGAN method, which uses random selection, has chances to select irrelevant patches (\ie the boy on the soccer court).
\section{Conclusions and Future Work}
\label{sec:conclusion}
\vspace{\secmargin}

In this work, we present a differentiable retrieval module to aid the image synthesis from the scene description.
Through the iterative process, the retrieval module selects mutually compatible patches as reference for the generation.
Moreover, the differentiable property enables the module to learn a better embedding function jointly with the image generation process.
Qualitative and quantitative evaluations validate that the synthesized images are realistic and diverse, while the retrieved patches are reasonable and compatible.

The proposed approach points out a new research direction in the content creation field.
As the retrieval module is differentiable, it can be trained with the generation or manipulation models to learn to select select real reference patches that improves the quality.
Future research in this area may include 1) improving the current patch pre-filtering approach to increase the number of candidate patches used in the retrieval; 2) designing better losses to facilitate the training of the retrieval process; 3) applying this technique to various conditional image/video generation or manipulation problems.

\section*{Acknowledgements}
\vspace{\secmargin}
We would like to thank the anonymous reviewers for their useful comments. This work is supported in part by the NSF CAREER Grant \#1149783.

\bibliographystyle{splncs04}
\bibliography{egbib}

\clearpage
\appendix
\section{Supplementary Materials}

\subsection{Overview}
In this supplementary material, we first show the critical steps in the weighted reservoir sampling approach.
We then present the implementation details of the proposed framework.
Finally, we supplement the training and evaluation details.

\subsection{Weighted Reservoir Sampling}
Our method is inspired by the weighted reservoir sampling approach in~\cite{xie2019reparameterizable}. The classical reservoir sampling~\cite{vitter1985random} is designed to sample $k$ items from a collection of n items~\cite{vitter1985random}. \cite{xie2019reparameterizable} shows this classical subset sampling procedure can be relaxed to a differentiable process by introducing 1) the Gumbel-max trick and 2) the relaxed top-$k$ function (Iterative Softmax). In Algorithm~\ref{alg:WRS}, we list the key steps using the notation defined in the main paper. IterSoftmax is a relaxed top-$k$ function that iteratively applies the Softmax operation (see Equation \eqref{eq:softmax}) to obtain the selection variable $s$.

\begin{algorithm}
\SetKwData{Left}{left}\SetKwData{This}{this}\SetKwData{Up}{up}
\SetKwInOut{Input}{Input}\SetKwInOut{Output}{Output}
\LinesNumbered
\Input{Patch bank $M$; subset size $n$; weights $\pi = [\pi_1, \ldots, \pi_{|M|}]$}
\Output{relaxed $n$-hot vector $s$ where $\sum_{i=1}^{|M|} s_i = n$ and $0 \le s_i \le 1$.}
\BlankLine
Initialize $\hat{\pi}$ as a zero vector of length $|M|$\;
\For{$i = 1, \ldots, |M|$ }{
    $u_i \leftarrow \text{Uniform}(0,1)$ \;
    $\hat{\pi}_i \leftarrow -\log(-\log(u_i)) + \log(\pi_i)$ \;
}
$s \leftarrow \text{IterSoftmax}(\hat{\pi}, n)$\;
\Return the relaxed $n$-hot vector $s$
\caption{Subset Selection using Weighted Reservoir Sampling.}
\label{alg:WRS}
\end{algorithm}

Our method is conceptually similar to Algorithm~\ref{alg:WRS}. But to make it work in our problem setting, we discuss two modifications in the main paper, including the group-wise sampling and the greedy selection strategy. The greedy selection strategy requires our model to update the query iteratively. Since the query is used to compute the weight $\hat{\pi}$, our algorithm presented in the main paper merges the iterative softmax with the Gumbel-max sampling step in Algorithm~\ref{alg:WRS}.

\subsection{Implementation Details}

\begin{figure}
    \centering
    \includegraphics[width=0.6\linewidth]{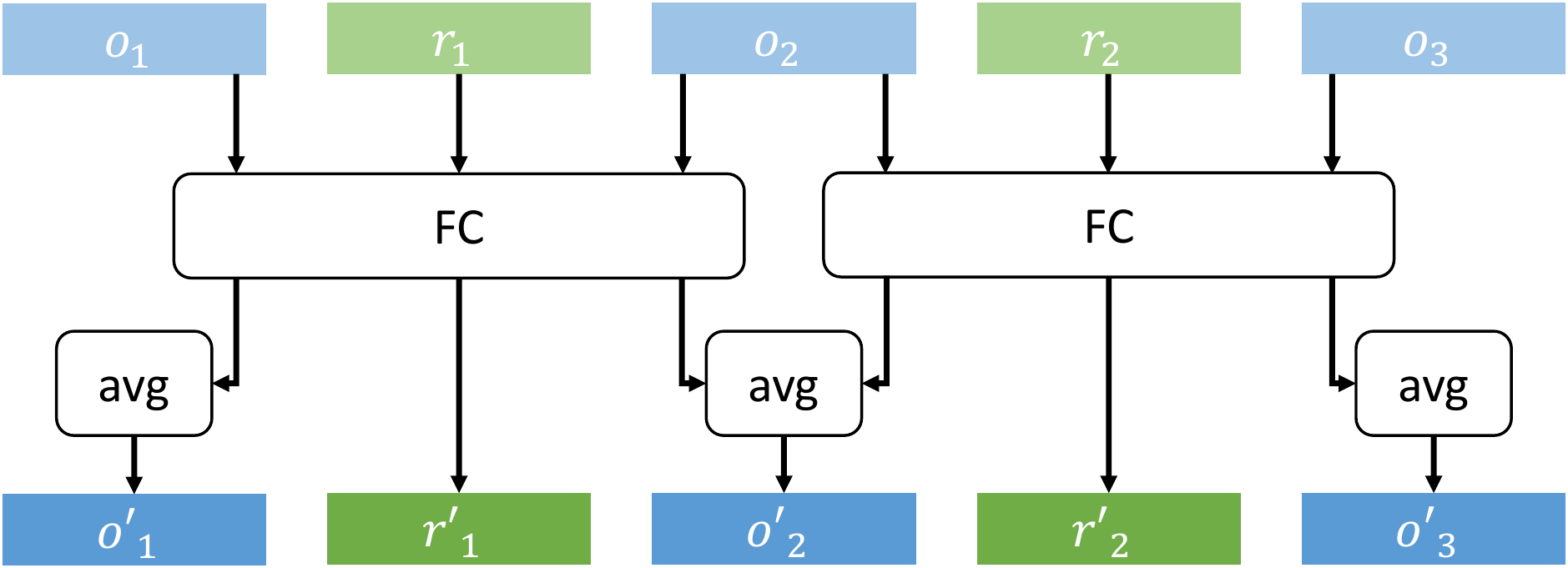}
    \vspace{-1mm}
    \caption{\textbf{Example of single graph convolutional layer.} Given an edge $e=(o_j,r_k,o_t)$, we first use a fully-connected layer to compute the output $o'_j, r'_k, o'_t$. We then apply the average pooling to fuse all the output vectors $o'_j$ computed from all the edges involving $o_j$ (\eg $o_2$ in this example).}
    \vspace{-3mm}
    \label{fig:gcn}
\end{figure}

\Paragraph{Scene graph encoder.}
We adopt the graph convolutional layers in sg2im~\cite{johnson2018image} for processing the input scene graph.
Specifically, given an edge $e=(o_j,r_k,o_t)$ in the scene graph, we compute the corresponding output vectors $o'_j$, $r'_k$, and $o'_t$ via a fully-connected layer.
The vectors $o_j$ and $o_t$ represent the object, while $r_k$ indicates the relation between $o_j$ and $o_t$.
However, since a single object $o_j$ may appear in multiple edges~(\ie participate in many relationships), we use the average pooling to fuse all the output vectors $o'_j$ computed from all the edges involving $o_j$.
We show an example of the single graph convolutional layer in~\figref{gcn}.
In practice, we use $5$ layers for our scene graph encoder $E$.

\Paragraph{Patch embedding function.}
The patch embedding function aims to compute the embedding of the candidate patches for the retrieval process.
We first use the pre-trained ResNet model~\cite{he2016deep} to extract the ImageNet features of all the patches in the patch memory bank.
We operate this process offline.
Then our patch embedding function is a series of fully-connected layers that maps the ImageNet feature space to the patch embedding space.

\Paragraph{Image Generation}
We use four modules to generate an image from a set of selected patches: crop encoder, object$^2$ refiner, object-image fuser, and decoder, described as follows:

\Paragraph{Crop encoder.}
The crop encoder extracts crop features $\{c_i\}$ from the selected patches.
We adopt a series of convolutional and down-sampling layers to build the crop encoder.

\Paragraph{Object$^2$ refiner.}
The object$^2$ refiner aims to associate the crop features with the relationships $\{r_k\}$ defined in the input scene graph.
Specifically, we replace the objects $\{o_i\}$ with the crop features $\{c_i\}$ in the edges defined in the input scene graph.
Similarly to the scene graph encoder, we use the graph convolutional layer to compute the output vectors $(c'_j, r'_k, c'_t)$ given the input edge $(c_j, r_k, c_t)$.
The average pooling function is then applied to combine the output vectors $c'_j$ computed from all the edges containing $c_j$.
Unlike the scene graph encoder, we use the 2D convolutional layer to build the graph convolutional layer since the input crop feature is of the dimension $D_c \times h \times w$.
We use $2$ layers in practice for the object$^2$ refiner.

\Paragraph{Object-image fuser.}
Given the refined object and predicate features, we use an object-image fuser to encode all features into a latent canvas $L$.
For each object, we first concatenate its refined crop features $c'_i$ and the original object feature $o_i$.
We then expand the concatenated feature to the shape of the corresponding predicted bounding box to get $u_i$ with dimension $D\times W \times H$.
Then we measure the attention map of each object by
\begin{equation}
    a_i = \frac{\exp(s_i)}{\sum_{j=i}^{N}\exp{(s_j)}}, 
\end{equation}
where $s_i = f(u_i)h(r_{p_i})$, $f$ and $h$ are learned mapping function, and $r_{p_i}$ is the relation feature for the relationship between the object and the image.
We can then obtain the final attention maps by summing up all object attention maps: $a = \sum_{i=1}^N a_i g(u_i)$, where $g$ is a learned mapping function.
Finally, we aggregate the attention maps to form the scene canvas with the `image' object features:
\begin{equation}
    L = \lambda_a a + u_{img},
\end{equation}
where $\lambda_a$ is the weight.

\Paragraph{Decoder.}
We use a series of convolutional and up-sampling layers to synthesize the final image from the scene canvas created by the object-image fuser.

\begin{table*}[t]
    \setlength{\tabcolsep}{4pt}
        \caption{\textbf{Statistics of datasets.} 
    }
    \label{tab:dataset}
    \centering
    \begin{tabular}{l c c c c c } 
	    \toprule
		Dataset & Train & Val & Test & $\#$ Category & $\#$ Patches \\
		\midrule
		COCO-Stuff & 74121 & 1024& 2048& 171& 411682\\
		Visual Genome & 62565& 5506 & 5088 & 178 & 606319\\
		\bottomrule
    \end{tabular}
\end{table*}
\begin{figure}[t]
    \centering
    \includegraphics[width=\linewidth]{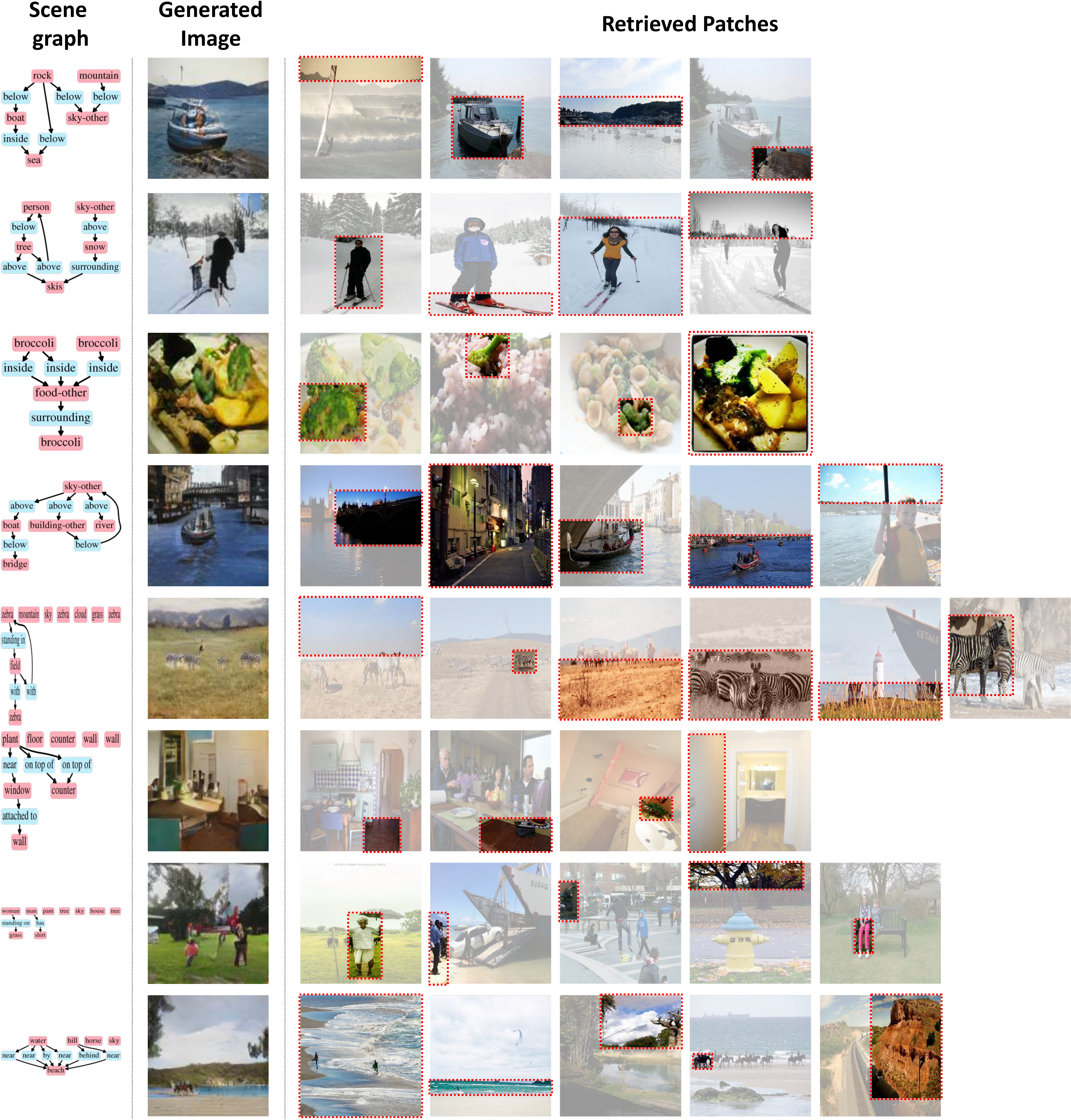}
    \vspace{-1.5mm}
    \caption{\textbf{Additional qualitative results.}
    For each sample, we show the retrieved patches which are used to guide the following image generation process. We also show the original image of each selected patch for clearer visualization.
    }
    \vspace{-1mm}
    \label{fig:retrieved_supp}
    \vspace{\figmargin}
    \vspace{-0mm}
\end{figure}

\section{Training and Evaluation}
\Paragraph{Training details.}
We implement with PyTorch~\cite{paszke2017automatic} and train our model with $90$ epochs on both the COCO-stuff~\cite{caesar2018coco} and visual genome~\cite{krishna2017visual} datasets.
We use the Adam optimizer~\cite{kingma2014adam} with a batch size of $16$.
The learning rates for the generator and discriminator are respectively $0.00025$ and $0.001$, and the exponential decay rates $(\beta_1,\beta_2)$ are set to be $(0,0.9)$.
We set the hyper-parameters as follows: $\lambda^\mathrm{sel}_\mathrm{gt}=0.1$, $\lambda^\mathrm{sel}_\mathrm{occur}=0.001$, $\lambda^\mathrm{img}_\mathrm{adv}=0.01$, $\lambda^\mathrm{img}_\mathrm{recon}=1$, $\lambda^\mathrm{img}_\mathrm{p}=1$, $\lambda^\mathrm{obj}_\mathrm{adv}=0.01$, $\lambda^\mathrm{obj}_\mathrm{ac}=0.1$, $\lambda^\mathrm{obj}_\mathrm{p}=0.5$, and $\lambda_\mathrm{bbx}L_\mathrm{bbx}=10$.

\Paragraph{Dataset summary.}
We present the summary of the datasets used for the training and evaluation in \tabref{dataset}.

\section{Additional Results}
We show more qualitative results of our approach in \figref{retrieved_supp}.
\end{document}